\pdfoutput=1
\documentclass[11pt]{article}

\usepackage[preprint]{acl}

\usepackage{times}
\usepackage{latexsym}
\usepackage{tabularray}
\usepackage[T1]{fontenc}
\usepackage{pdflscape} % put this in your preamble

\usepackage[utf8]{inputenc}

\usepackage{microtype}

\usepackage{inconsolata}

\usepackage{graphicx}
\usepackage{amsmath}
\title{Multi-Dimensional Prompt Chaining to Improve Open-Domain Dialogue Generation}

\author{Livia Leong Hui Teng\textsuperscript{1}, Jing Yang Lee\textsuperscript{2}\\
  Nanyang Technological University \\
  \texttt{\{lleong013\textsuperscript{1}, jingyang001\textsuperscript{2}\}@e.ntu.edu.sg} \\}

\begin{document}
\maketitle
\begin{abstract}
Small language models (SLMs) offer significant deployment advantages but often struggle to match the dialogue quality of larger models in open-domain settings. In this paper, we propose a multi-dimensional prompt-chaining framework that integrates Naturalness, Coherence, and Engagingness dimensions to enhance human-likeness in open-domain dialogue generation. We apply the framework to two SLMs—TinyLlama and Llama-2-7B—and benchmark their performance against responses generated by substantially larger models, including Llama-2-70B and GPT-3.5 Turbo. We then employ automatic and human evaluation to assess the responses based on diversity, contextual coherence, as well as overall quality. Results show that the full framework improves response diversity by up to 29\%, contextual coherence by up to 28\%, and engagingness as well as naturalness by up to 29\%. Notably, Llama-2-7B achieves performance comparable to substantially larger models, including Llama-2-70B and GPT-3.5 Turbo. Overall, the findings demonstrate that carefully designed prompt-based strategies provide an effective and resource-efficient pathway to improving open-domain dialogue quality in SLMs.
\end{abstract}

\section{Introduction}
Large language models (LLMs) have revolutionized natural language processing, demonstrating remarkable capabilities in understanding context and generating human-like responses \citep{devlin_bert_2019}. These advances in open-domain dialogue generation enable more meaningful and engaging conversations with users, with promising applications ranging from enhanced user engagement to mental health support \citep{siddals_it_2024}. Recent researches has focused on improving response quality through various approaches, including generating more diverse responses \citep{liu_promoting_2023,lee2023empirical, 10.1609/aaai.v37i11.26594}, adapting flexible strategies or frameworks \citep{9966598, wang2022advancedconditionalvariationalautoencoders}, incorporating social norms and expressions \citep{varshney_art_2024, 10.1145/3806774.3827985}, or improving multi-turn performance \citep{laban2025llmslostmultiturnconversation, lee-etal-2025-doctalk, yi2025surveyrecentadvancesllmbased}.

However, these advances are predominantly confined to large-scale models that require substantial computational resources to operate efficiently. In contrast, Small Language Models (SLMs) offer significant advantages in terms of computational efficiency, cost-effectiveness, and adaptability \citep{wang_comprehensive_2024}, but struggle to achieve comparable dialogue quality.  To bridge this performance gap between LLMs and SLMs, prompt-based techniques - especially few-shot in-context learning - has emerged as a promising approach for enhancing model performance without additional training or modifying model parameters, with demonstrated effectiveness for both LLMs \citep{brown_language_2020} and SLMs \citep{schick_its_2021}.

Hence, in this paper, we introduce a novel multidimensional prompt chaining framework that enables SLMs to achieve performance comparable to larger models in open-domain dialogue generation. Prompt chaining decomposes complex tasks into sequential subtasks, where intermediate outputs from one prompt feed into subsequent prompts. Our framework leverages this approach to iteratively refine generated responses through a structured chain in which each prompt focuses on enhancing a distinct dimension of response quality, specifically contextual coherence, naturalness, and engagingness. Through systematic experimentation with various few-shot learning configurations, we provide empirical evidence that our approach significantly enhances response quality across both quantitative metrics and qualitative assessments, enabling SLMs to perform on par with substantially larger and more resource-intensive LLMs.

The remainder of this paper is organized as follows: We first present our methodology, including the few-shot generation approach and response generation workflow. We then describe our experimental variations and evaluation metrics, followed by automatic metrics and human evaluation results and discussion of our findings.

\section{Methodology}
In this paper, we propose an In-Context Learning prompt chaining framework to improve the coherence, engagingness and naturalness of open-domain dialogue responses. We selected these three dimensions to prioritize human-likeness in open-ended conversational settings \citep{zhong2022unifiedmultidimensionalevaluatortext, finch-choi-2020-towards,gopalakrishnan_topical-chat_2019}. The quality of the performance is then evaluated, typically based on multiple dimensions of the response, namely, coherence, engagement and naturalness.

The framework iterates refinement based on specific qualitative criteria, as illustrated/outlined below. 
\begin{figure}[h]
\scalebox{0.4}{
\centering\includegraphics[]{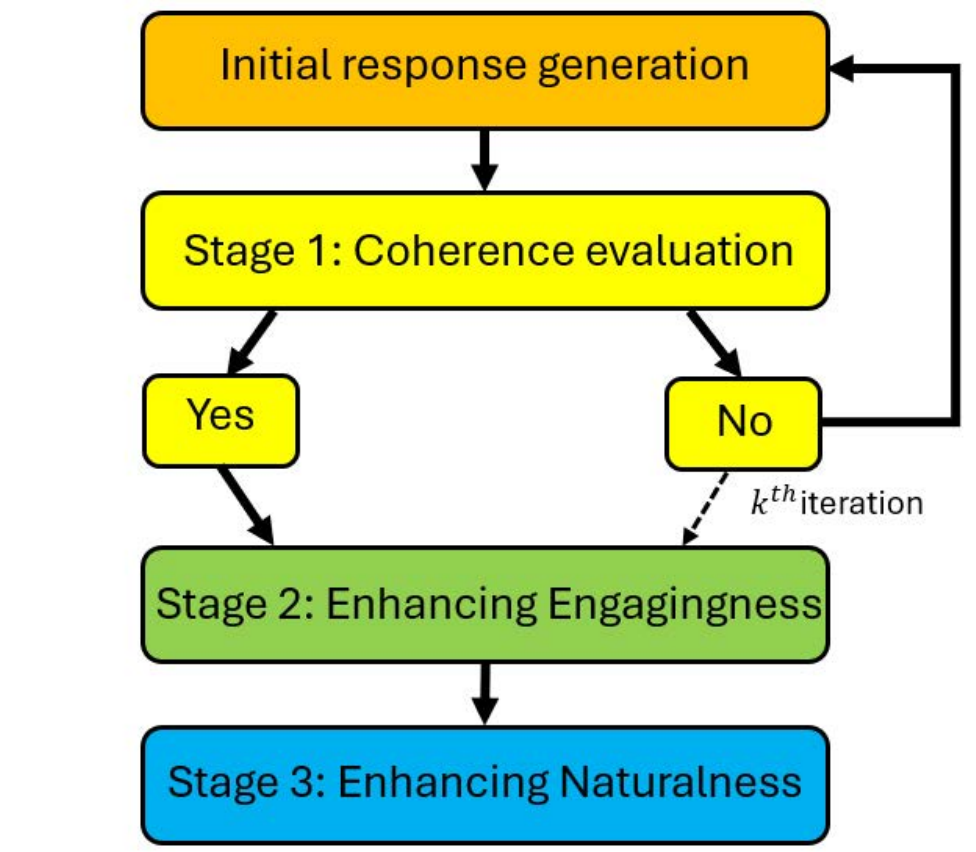}}
\caption{\label{fig:1} Workflow of the response generation framework. The process includes: 
    initial response generation, (1) coherence evaluation with up to $k$ iterations, 
    (2) engagingness improvement if coherence is achieved and (3) naturalness improvement to finalize the response.}
\end{figure}
\subsection{Initial Response Generation}
\label{subsubsec:IRG}
The first response is generated using a zero-shot approach, with the utterance–response dialogue history provided as input to the SLM. The model is instructed to adopt the speaker's persona and continue the conversation based on the preceding context. This setup enables the generation of contextually coherent responses without requiring explicit demonstration samples.

\begin{figure*}
    \centering
    \includegraphics[width=0.75\linewidth]{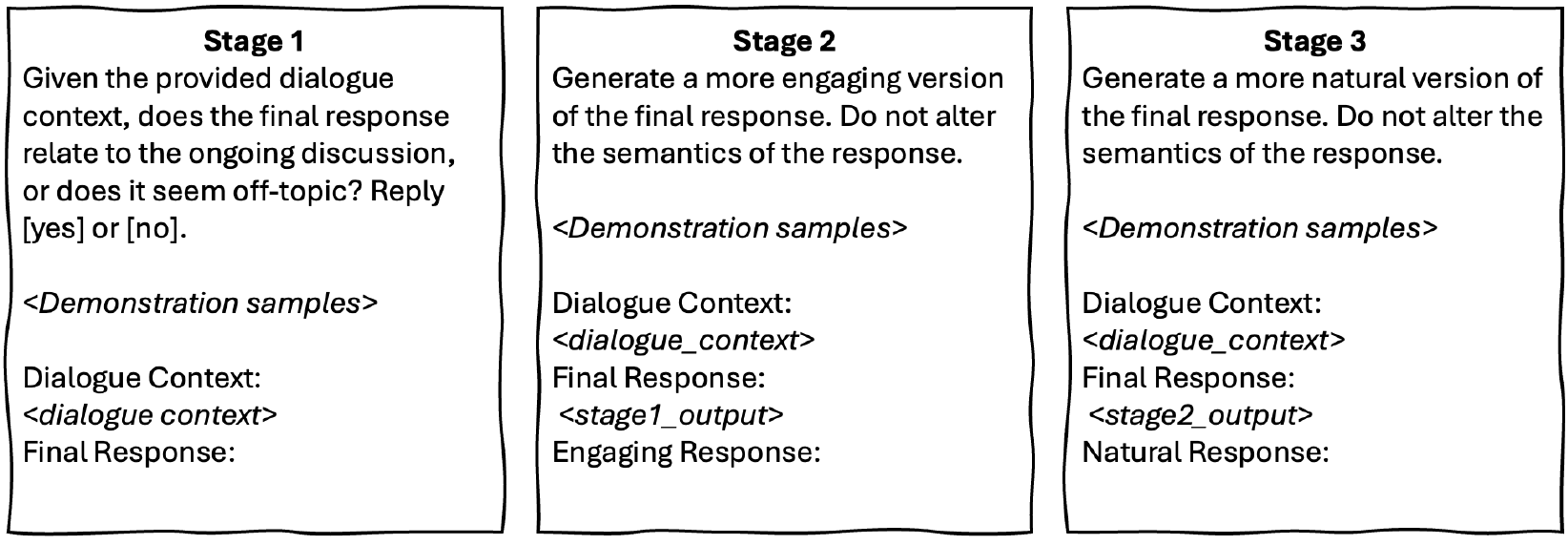}
    \caption{Prompt templates for Stage 1,2 and 3 of the pipeline.}
    \label{fig:prompt_templates}
\end{figure*}

\subsection{Stage 1: Coherence Evaluation}    
The first stage evaluates whether the generated response is contextually coherent with respect to the dialogue history. Coherence is required to maintain good conversational flow by being consistent and minimizing repetition, disfluency and semantic errors \citep{see_what_2019,shin_generating_2021}. This evaluation employs a three-shot in-context learning prompt, where each demonstration comprises a dialogue context, a reference response, and a randomly selected utterance from a separate conversation. The dialogue context paired with its reference response serves as a positive example, while the utterance from the unrelated conversation serves as a negative example.

To construct these demonstrations, we leverage the training set of the DailyDialog dataset, which provides the dialogue context and reference response for each sample. Using an LLM, we generated incoherent responses for each dialogue context to serve as negative counterparts. Both the reference and incoherent responses were scored using UniEval, a top-performing unified evaluator that employs a question-answering framework to assess multiple dimensions of text generation quality, including coherence, engagingness, and naturalness \citep{zhong2022unifiedmultidimensionalevaluatortext}. UniEval Coherence scores are used to select positive and negative demonstrations, corresponding to the highest and lowest-scoring context-response pairs respectively.

Following these demonstrations, the model is tasked with classifying its own response as coherent ("Yes") or incoherent ("No"). If the response is deemed incoherent, the process returns to the initial generation stage (Section \ref{subsubsec:IRG}) to produce a new response. This loop terminates once either a contextually coherent response is generated or $k$ iterations have been reached. In our evaluation, the iteration limit $k$ was set to 5 as preliminary experiments indicated diminishing returns beyond this threshold.
    
\subsection{Stage 2: Enhancing Engagingness} 
If the response is contextually coherent, the SLM is prompted to revise the response to enhance its engagingness. Engagement ensures that the chatbot’s response is novel while encouraging the conversation to continue \citep{yi_towards_2019}. This stage employs a three-shot prompt, with demonstrations drawn from the DailyDialog training set. For this stage, we generate unengaging responses for each dialogue context by explicitly prompting an LLM to generate laconic and passive responses. Both these generated responses and the reference responses were then evaluated using UniEval's engagingness dimension:
\begin{equation}
\text{Diff}_{\text{eng}} = 
S_{\text{ref}}^{\text{E}} - S_{\text{uneng}}^{\text{E}}
\end{equation}
where $S_{\text{ref}}^{\text{E}}$ and $S_{\text{uneng}}^{\text{E}}$ refer to the UniEval engaginess score of the reference response and the unengaging response respectively. The three dialogues exhibiting the highest $\text{Diff}_{\text{eng}}$ values are used as demonstrations, with the unengaging responses presented as negative examples and the corresponding reference responses as positive examples of engaging output.

\subsection{Stage 3: Enhancing Naturalness} 
In Stage 3, the SLM is prompted to improve the naturalness of the response.  Naturalness draws the distinction between the phrasing of the response, targeting enhanced conversational flow and more human-like expression \citep{see_what_2019,zhang_dialogpt_2020}. This stage also follows a three-shot approach, again utilizing the DailyDialog training set. Similarly, to generate a pool of demonstration samples, we explicitly prompt an LLM to generate unnatural responses for each dialogue context, and scored both these responses and the reference responses using UniEval's naturalness dimension. Demonstrations are selected based on the largest differences between reference and unnatural response scores:
\begin{equation}
\text{Diff}_{\text{nat}} =
S_{\text{ref}}^{\text{N}} - S_{\text{unnat}}^{\text{N}}
\end{equation}
where $S_{\text{ref}}^{\text{N}}$ and $S_{\text{uneng}}^{\text{N}}$ refer to the UniEval naturalness score of the reference response and the unnatural response respectively.

\section{Experimental Design}
We evaluate our proposed framework on TinyLlama \cite{zhang2024tinyllamaopensourcesmalllanguage} and the chat variant of Llama-2-7B \cite{touvron2023llama2openfoundation}. Dialogue contexts are sourced from the DailyDialog dataset obtained from HuggingFace, which consists of multi-turn open-domain conversations that reflect human daily communication without predefined roles or knowledge grounding \citep{li_dailydialog_2017}. The dataset is human-annotated and captures natural expressions and emotions, making it an ideal benchmark for evaluating conversational quality in naturalistic settings. We compare the results with the language models' unprompted baseline generated responses, and, ablate prompt variations to identify the significance of each dimensions. 

\subsection{Ablation Study}
To investigate the contribution of each dimension to the improved overall quality, we tested 4 configurations of the framework in addition to the base SLM. 
\begin{enumerate}
    \item \textbf{Full framework:} Full pipeline.
    \item \textbf{w/o coherence:} Only Stage 2 and 3 of the pipeline.
    \item \textbf{w/o engagingness:} Only Stage 1 and 2 of the pipeline.
    \item \textbf{w/o naturalness:} Only Stage 1 and 3 of the pipeline.
    \item \textbf{Base:} Directly prompting the base SLM without applying our pipeline.
\end{enumerate}
Each combination was applied for the same dialogue context and response to generate four responses for each set of dialogue context. Additionally, we also benchmark our approach by evaluating responses generated by directly prompting the chat variant of Llama2-70b and gpt=3.5-turbo.

\subsection{Evaluation Metrics}
To assess the quality of the generated responses across all configurations, we employed three evaluation metrics to ensure robust statistical analysis.
\begin{enumerate}
    \item \textbf{UniEval:} Using the UniEval LLM-as-a-judge framework \cite{zhong2022unifiedmultidimensionalevaluatortext}, we extracted scores for coherence, engagingness, and naturalness.
    
    \item \textbf{Utterance Entailment (UE) score:} Metric that quantifies contextual coherence by computing the Natural Language Inference score between the generated response and each utterance in the dialogue context \citep{9747458}.
    \item \textbf{Distinct-N:} A diversity metric that measures the proportion of unique n-grams, which we applied unigrams, bigrams and trigrams in generated responses \citep{li_diversity-promoting_2016}. Higher values indicate more varied and less repetitive outputs.
\end{enumerate}

We normalized the scores where necessary to allow for a fair comparison across the metrics.

\section{Results and Discussion}

\begin{table*}
\centering
\caption{Quantitative Evaluation Scores. For Tinyllama and Llama-2-7B, the full prompt framework, and ablations without each individual dimensions, and the base response are evaluated. Scores are derived from Distinct-1, 2 and 3, UniEval's individual dimensional scores, and the UE scores.}
\label{tab:fullscores}
\begin{tblr}{
  row{2} = {c},
  cell{1}{2} = {c},
  cell{1}{3} = {c},
  cell{1}{4} = {c},
  cell{1}{5} = {c},
  cell{1}{6} = {c},
  cell{1}{7} = {c},
  cell{1}{8} = {c},
  cell{3}{2} = {c},
  cell{3}{3} = {c},
  cell{3}{4} = {c},
  cell{3}{5} = {c},
  cell{3}{6} = {c},
  cell{3}{7} = {c},
  cell{3}{8} = {c},
  cell{4}{2} = {c},
  cell{4}{3} = {c},
  cell{4}{4} = {c},
  cell{4}{5} = {c},
  cell{4}{6} = {c},
  cell{4}{7} = {c},
  cell{4}{8} = {c},
  cell{5}{2} = {c},
  cell{5}{3} = {c},
  cell{5}{4} = {c},
  cell{5}{5} = {c},
  cell{5}{6} = {c},
  cell{5}{7} = {c},
  cell{5}{8} = {c},
  cell{6}{2} = {c},
  cell{6}{3} = {c},
  cell{6}{4} = {c},
  cell{6}{5} = {c},
  cell{6}{6} = {c},
  cell{6}{7} = {c},
  cell{6}{8} = {c},
  cell{7}{2} = {c},
  cell{7}{3} = {c},
  cell{7}{4} = {c},
  cell{7}{5} = {c},
  cell{7}{6} = {c},
  cell{7}{7} = {c},
  cell{7}{8} = {c},
  cell{8}{1} = {c},
  cell{9}{2} = {c},
  cell{9}{3} = {c},
  cell{9}{4} = {c},
  cell{9}{5} = {c},
  cell{9}{6} = {c},
  cell{9}{7} = {c},
  cell{9}{8} = {c},
  cell{10}{2} = {c},
  cell{10}{3} = {c},
  cell{10}{4} = {c},
  cell{10}{5} = {c},
  cell{10}{6} = {c},
  cell{10}{7} = {c},
  cell{10}{8} = {c},
  cell{11}{2} = {c},
  cell{11}{3} = {c},
  cell{11}{4} = {c},
  cell{11}{5} = {c},
  cell{11}{6} = {c},
  cell{11}{7} = {c},
  cell{11}{8} = {c},
  cell{12}{2} = {c},
  cell{12}{3} = {c},
  cell{12}{4} = {c},
  cell{12}{5} = {c},
  cell{12}{6} = {c},
  cell{12}{7} = {c},
  cell{12}{8} = {c},
  cell{13}{2} = {c},
  cell{13}{3} = {c},
  cell{13}{4} = {c},
  cell{13}{5} = {c},
  cell{13}{6} = {c},
  cell{13}{7} = {c},
  cell{13}{8} = {c},
  cell{14}{2} = {c},
  cell{14}{3} = {c},
  cell{14}{4} = {c},
  cell{14}{5} = {c},
  cell{14}{6} = {c},
  cell{14}{7} = {c},
  cell{14}{8} = {c},
  cell{15}{2} = {c},
  cell{15}{3} = {c},
  cell{15}{4} = {c},
  cell{15}{5} = {c},
  cell{15}{6} = {c},
  cell{15}{7} = {c},
  cell{15}{8} = {c},
  hline{1-3,8-9,14,16} = {-}{},
  stretch=0, colsep  = 3.0pt, rowsep=2pt
}
                    & \textbf{Dist-1} & \textbf{Dist-2} & \textbf{Dist-3} & {\textbf{UniEval -}\\\textbf{ Naturalness}} & {\textbf{UniEval -}\\\textbf{ Coherence}} & {\textbf{UniEval -}\\\textbf{ Engagingness}} & \textbf{UE} \\
\textbf{Tinyllama}  &                 &                 &                 &                                             &                                           &                                              &             \\
Full                & 0.28            & 0.71            & 0.86            & 0.81                                        & 0.84                                      & 2.16                                         & 0.28        \\
w/o Coherence       & 0.26            & 0.73            & 0.89            & 0.72                                        & 0.72                                      & 2.02                                         & 0.22        \\
w/o Naturalness     & 0.26            & 0.65            & 0.83            & 0.66                                        & 0.75                                      & 2.21                                         & 0.25        \\
w/o Engagingness    & 0.25            & 0.72            & 0.78            & 0.69                                        & 0.73                                      & 1.56                                         & 0.24        \\
Base                & 0.25            & 0.55            & 0.82            & 0.63                                        & 0.7                                       & 1.99                                         & 0.2         \\
\textbf{Llama-2 7B} &                 &                 &                 &                                             &                                           &                                              &             \\
Full                & 0.32            & 0.79            & 0.91            & 0.88                                        & 0.89                                      & 2.45                                         & 0.32        \\
w/o Coherence       & 0.27            & 0.74            & 0.86            & 0.83                                        & 0.77                                      & 2.17                                         & 0.25        \\
w/o Naturalness     & 0.29            & 0.7             & 0.85            & 0.75                                        & 0.8                                       & 2.22                                         & 0.27        \\
w/o Engagingness    & 0.22            & 0.72            & 0.77            & 0.79                                        & 0.81                                      & 1.87                                         & 0.25        \\
Base                & 0.29            & 0.62            & 0.83            & 0.7                                         & 0.78                                      & 2.07                                         & 0.22        \\
\textbf{Llama-2-70b}         & 0.30            & 0.77            & 0.88            & 0.86                                        & 0.87                                      & 2.33                                         & 0.28        \\
\textbf{gpt-3.5-turbo}       & 0.31            & 0.79            & 0.92            & 0.87                                        & 0.92                                      & 2.39                                         & 0.31        
\end{tblr}
\end{table*}
\begin{table}
\centering
\caption{Quantitative Human Evaluation. Similar to \citet{shi_novel_2023-1, lee_modeling_2025-1}, we engage 5 annotators to assess the overall quality of responses generated by Llama2-7b (using our full pipeline) against those of Llama2-70b and GPT-3.5-Turbo. The `Win', `Tie', and `Loss' percentages indicate the proportion of Llama2-7b-generated responses deemed to be of lower quality, comparable quality, or better quality, respectively, relative to Llama2-70b and GPT-3.5-Turbo.}
\label{tab:humanscores}
\begin{tblr}{
  column{even} = {c},
  column{3} = {c},
  hline{1-2,5} = {-}{},
  stretch=0, colsep  = 3.0pt, rowsep=2pt
}
                               & \textbf{Win} & \textbf{Tie} & \textbf{Loss} \\
\textbf{Full vs Base}          & 59\%         & 22\%         & 19\%          \\
\textbf{Full vs Llama2-70b}    & 34\%         & 42\%         & 24\%          \\
\textbf{Full vs gpt-3.5-turbo} & 33\%         & 35\%         & 32\%          
\end{tblr}
\end{table}

Quantitative human evaluation shows that the \textbf{full framework in Llama 2-7B achieves results comparable to Llama 2-70B and GPT-3.5}. The full framework achieves the \textbf{highest lexical diversity} for both TinyLlama and Llama-2-7B, with Distinct-1 scores outperforming their respective baselines by 0.03. This indicates a broader vocabulary in responses, enhancing engagement through varied word choice. The full framework also yields \textbf{high phrase diversity} for Distinct-2 and 3, with TinyLlama scoring 0.71 (Distinct-2) and 0.86 (Distinct-3) compared to baseline 0.55 and 0.82, respectively, and Llama-2 7B scoring 0.79 and 0.91 against baseline 0.62 and 0.83. These gains reflect stronger diversity in multi-word sequences, supporting engaging and less repetitive dialogues when the full framework is applied. More natural and engaging dialogues lead to stronger phrase diversity as they force the model to use a wider vocabulary, generating responses that are less cursory and less generic.

Individual dimension scoring and ablation studies reveal a degree of \textbf{Naturalness-Engagingness interdependence}. For Tinyllama, the UniEval-Engagingness scored the highest (2.21) when Naturalness prompt is excluded, suggesting that Naturalness constraint may over-regularize or limit the linguistic creativity. While Naturalness scoring favours conventional and grammatically neutral phrasing, engaging responses often rely on expressiveness, emotional tone, and stylistic variation. Enforcing the Naturalness component may unintentionally bias Tinyllama toward safer, more formulaic outputs—thereby dampening its engaging qualities. However, the interdependence trend is not reflected in Llama-2-7B. Although the ablation without Naturalness still has a high UniEval-Engagingness score (2.22), the full framework remains the highest scoring for UniEval-Engagingness (2.45). Llama-2-7B may intrinsically generate more expressive or stylized text. The Naturalness requirement helps refine and stabilize that expressiveness. In this case, Naturalness and Engagingness are complementary rather than competing components. Overall, the interaction between Naturalness and Engagingness is model-dependent, functioning as competing objectives in smaller models but complementary dimensions in larger ones.

Coherence introduces a mild trade-off, modestly constraining Naturalness and Engagingness while remaining critical to overall response quality. \textbf{Coherence plays a stabilising role}, with its removal allowing greater stylistic freedom and expressiveness. Nevertheless, the full framework consistently achieves the highest scores across all dimensions, indicating that Coherence remains essential in maintaining structural clarity even if it slightly constraints creativity. This is further supported by the highest UE-scoring full framework response, reduced dimension ablations reduced UE scores. Overall, these findings indicate that Coherence contributes to precision and structure, with subtle creativity trade-offs in expressiveness.

Overall, the full pipeline produces more diverse, natural, coherent, and engaging responses, achieving significantly greater scores on all automatic metrics. Human evaluations corroborate these quantitative results, consistently rating outputs from the full framework as superior to those from the base SLM. Additionally, when the full pipeline is applied, responses generated by SLMs are generally comparable to those generated by much larger counterparts such as Llama2-70b and gpt-3.5-turbo in terms of both automatic metrics and human evaluation. Notably, when compared against Llama-2-70B, applying our pipeline to Llama2-7b yields even better performance, effectively narrowing the quality gap between SLMs and LLMs.

\section{Related Work}

In-context learning, pioneered by GPT-3 \citep{brown_language_2020}, has emerged as a powerful paradigm that enables language models to adapt to new tasks by conditioning on a few demonstration examples within the input prompt, without requiring parameter updates. In recent years, researchers have developed more sophisticated techniques including chain-of-thought prompting \citep{wei_chain--thought_2023}, tree-of-thought prompting \citep{yao_tree_2023}, and self-consistency prompting\citep{wang_self-consistency_2023}. In the context of dialogue generation, in-context learning has shown promise. Recent studies have applied few-shot prompting to enhance dialogue systems, demonstrating improvements in empathetic response generation \citep{cai_empcrl_2024}, information-seeking dialogue \citep{lee_redefining_2024}, and persona-consistent dialogue \citep{xu_towards_2023}. With regard to open-domain dialogue specifically, prior work have leveraged in-context learning to learn implicit pattern information between contexts and responses \citep{liu_promoting_2023}, model the one-to-many relationship \citep{lee_redefining_2024}, and to generated relevant questions in mixed initiative open-domain conversations\citep{ling_generating_2023}.

%Open-domain dialogue generation has evolved from early rule-based systems such as ELIZA \citep{weizenbaum1966eliza} and PARRY \citep{colby1971artificial}, which relied on predefined scripted pattern matching, to retrieval-based methods that drew responses from existing repositories, but often produced repetitive or mismatched output \citep{Yan 2016}. The introduction of sequence-to-sequence learning shifted generative paradigms, providing an end-to-end response in dialog generation, which formed the foundation of modern LLM-driven conversational models \citep{Sutskever 2014}. However, smaller LLMs still face performative constraints \citep{Shen 2024}, , prompting the rise of lightweight strategies such as prompt engineering and few-shot prompting \citep{brown_language_2020} \citep{qiao 2024}, which guide model behaviour without parameter updates and offer a practical alternative to more resource-intensive fine-tuning \citep{wang_comprehensive_2024}.

\section{Conclusion} 
This study shows that integrating Naturalness, Coherence, and Engagingness within a multi-dimensional prompt-chaining framework significantly improves the response quality of smaller language models. The full framework consistently enhances lexical and phrasal diversity, producing more natural, coherent, and engaging dialogue, with both automatic metrics and human evaluations indicating increased human-likeness. Ablation results reveal model-dependent trade-offs between expressiveness and structure, but demonstrate that combining all dimensions yields the most balanced outputs. Overall, these findings highlight that our approach offers a practical and resource-efficient pathway for narrowing the quality gap between SLMs and larger LLMs in open-domain dialogue generation. Future work could explore refined prompt engineering at each stage of the framework, as well as supervised fine-tuning approaches, to further close the performance gap between SLMs and their larger counterparts.

\bibliography{citations.bib}
\clearpage

% \appendix

% \section{Example Appendix}
% \label{sec:appendix}

\end{document}